\documentclass{article}
\usepackage{spconf,amsmath,graphicx,hyperref}
\usepackage{booktabs}
\usepackage{caption}

\title{Vi-SAFE: A Spatial\mbox{-}Temporal Framework for Efficient Violence Detection in Public Surveillance}
\name{Ligang Chang$^{1}$, Shengkai Xu$^{3}$,Liangchang Shen$^{2}$,Binhan Xu$^{1}$,Junqiao Wang$^{3}$, Tianyu Shi$^{4}$,Yanhui Du$^{1*}$\thanks{$^*$Corresponding author: duyanhui@ppsuc.edu.cn}}
\address{$^{1}$School of Information Network Security, People's Public Security University of China, Beijing, China\\
$^{2}$Department of Architecture and Civil Engineering, City University of Hong Kong, Hong Kong, China\\
$^{3}$College of Computer Science, Sichuan University, Chengdu, China\\
$^{4}$Faculty of Applied Science and Engineering, University of Toronto, Canada
}

\begin{document}
%
\maketitle
\begin{abstract}

The automatic detection of violent behaviors, such as physical altercations in public areas, is critical for public safety. This study addresses challenges in violence detection, including small-scale targets, complex environments, and real-time temporal analysis. We propose Vi-SAFE, a spatial–temporal framework that integrates an enhanced YOLOv8 with a Temporal Segment Network (TSN) for video surveillance. The YOLOv8 model is optimized with GhostNetV3 as a lightweight backbone, an exponential moving average (EMA) attention mechanism, and pruning to reduce computational cost while maintaining accuracy. YOLOv8 and TSN are trained separately on pedestrian and violence datasets, where YOLOv8 extracts human regions and TSN performs binary classification of violent behavior. Experiments on the RWF-2000 dataset show that Vi-SAFE achieves an accuracy of 0.88, surpassing TSN alone (0.77) and outperforming existing methods in both accuracy and efficiency, demonstrating its effectiveness for public safety surveillance.Code is available at \url{https://anonymous.4open.science/r/Vi-SAFE-3B42/README.md}.
\end{abstract}
\section{Introduction}
\label{sec:intro}

Accelerating urbanization and rising social complexity have heightened the demand for public safety monitoring as violent incidents, such as for public safety altercations and physical confrontations, continue to increase \cite{1}. These events threaten individuals and communities, cause property damage, and their unpredictability hampers timely intervention. As a result, research emphasizes real-time detection and intervention in high-risk public settings, including transportation hubs, schools, and urban surveillance systems \cite{2}.

Violence recognition in video surveillance has gained growing attention. Early approaches relied on 2D CNNs, effective in spatial analysis but limited in temporal modeling, while 3D CNNs and hybrid architectures improved accuracy at the cost of high computational demand \cite{5, 21}. Methods such as TSN \cite{15} and Two-Stream Networks \cite{18} partially addressed temporal dependencies but remain inefficient for edge deployment. More recent models, including CNN–LSTM hybrids \cite{5, 8} and dual-stream frameworks \cite{12, 14}, enhance spatiotemporal features yet still struggle with real-time performance. Object detectors like YOLO \cite{36, 37} offer efficiency but lack robust temporal reasoning, highlighting the need for approaches that balance accuracy and efficiency for low-power edge devices.

This study introduces Vi-SAFE, a spatial–temporal framework for violence detection that integrates an optimized YOLOv8s object detector with a temporal segment network (TSN) \cite{15}. The framework aims to balance accuracy and efficiency, making it suitable for deployment on resource-constrained edge devices. In Vi-SAFE, YOLOv8s is enhanced with GhostNetV3 \cite{16} as a lightweight backbone to reduce complexity and an exponential moving average (EMA) attention mechanism \cite{17} to strengthen feature extraction. Pruning techniques further compress the model, lowering parameters and FLOPs while preserving accuracy. After detecting regions of interest (ROIs) related to potential violent actions, TSN performs temporal analysis to capture motion dynamics and suppress background interference. This integration improves recognition accuracy, efficiency, and robustness in complex surveillance scenarios. The main contributions are as follows:

\subsection{Lightweight detection with enhanced features}
YOLOv8s is optimized by integrating GhostNetV3, EMA attention, and pruning, achieving better accuracy–efficiency trade-offs and enabling deployment on edge devices.

\subsection{Temporal reasoning for violence recognition}
A lightweight TSN models temporal dynamics within human ROIs, improving recognition of violent behaviors while reducing background interference.

\subsection{Unified spatial–temporal framework with validated effectiveness}
Vi-SAFE provides an end-to-end framework that combines efficient object detection with temporal reasoning. Experiments on RWF-2000 confirm that it outperforms mainstream approaches in both accuracy and efficiency, demonstrating its practicality for public safety surveillance.

\section{Methods}
\label{sec:methods}

\begin{figure}[htb]
\centering
\includegraphics[width=\linewidth]{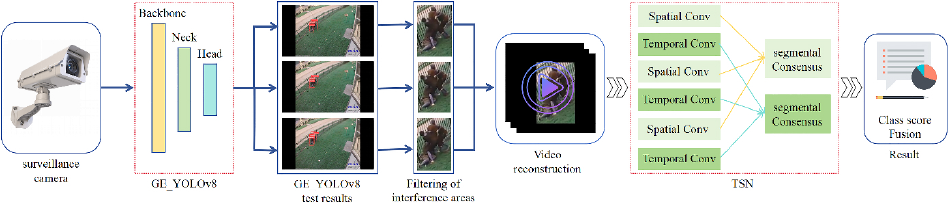}
\caption{Overview of the Vi-SAFE framework.}
\label{fig:fig1}
\end{figure}

We propose Vi-SAFE, a spatial–temporal framework for violence detection that integrates the GE-YOLOv8 object detector with a temporal segment network (TSN). In each video frame, GE-YOLOv8 focuses on accurately localizing individuals potentially involved in violent activities, while the TSN models the temporal dynamics of these regions of interest (ROIs) to determine violent behavior. The overall architecture of Vi-SAFE is illustrated in Fig\ref{fig:fig1}.

\subsection{GE\mbox{-}YOLOv8 model}
\label{ssec:subhead7}
\begin{figure}[htb]
\centering
\includegraphics[width=0.9\linewidth]{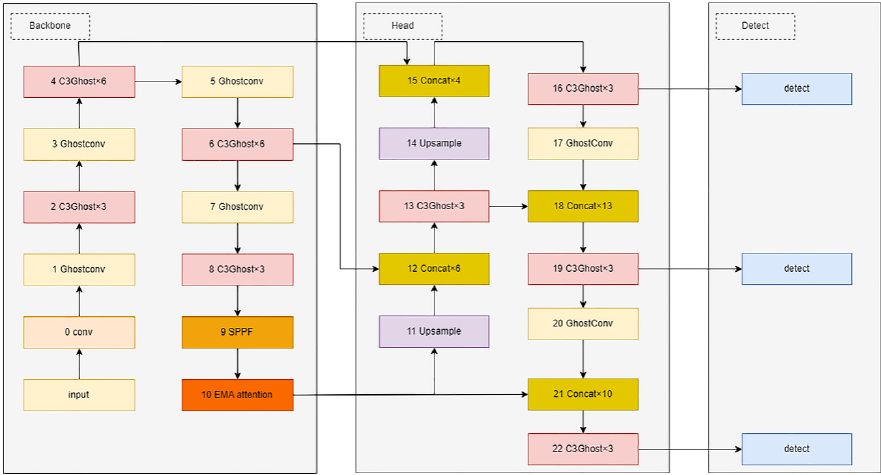}
\caption{Network structure based on GE\mbox{-}YOLOv8 configuration.}
\label{fig:fig2}
\end{figure}

YOLOv8 has five variants (n, s, m, l, x) with increasing width and depth \cite{39}. For a balance between accuracy and efficiency, YOLOv8s is chosen as the baseline. The backbone and head are replaced with GhostNetV3 \cite{16} to reduce redundancy, and an EMA attention mechanism \cite{17} is added to strengthen feature extraction in complex scenes. Structured pruning further compresses parameters and FLOPs while maintaining accuracy. The optimized model is referred to as ``GE\mbox{-}YOLOv8,'' whose structure is shown in Fig.~\ref{fig:fig2}.

\subsubsection{GhostNetV3 backbone}
\label{sssec:subsubhead1}
\begin{figure}[htb]
\centering
\includegraphics[width=.72\linewidth]{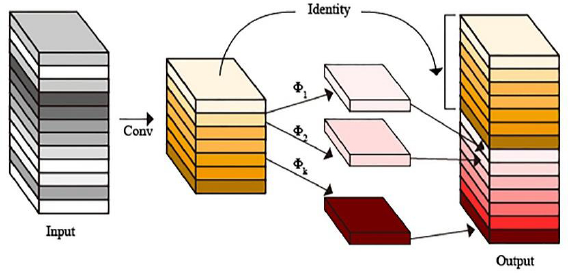}
\caption{GhostConv architecture in GhostNetV3.}
\label{fig:fig3}
\end{figure}

The backbone of YOLOv8s is replaced with GhostNetV3 \cite{16}, a lightweight CNN that generates additional feature maps through inexpensive linear operations instead of redundant convolutions. GhostConv and C3Ghost modules are integrated into both backbone and head (Fig.~\ref{fig:fig3}), which substantially reduce FLOPs and parameters while maintaining accurate feature representation.

\subsubsection{EMA attention mechanism}
\label{sssec:subsubhead2}
\begin{figure}[htb]
\centering
\includegraphics[width=0.9\linewidth]{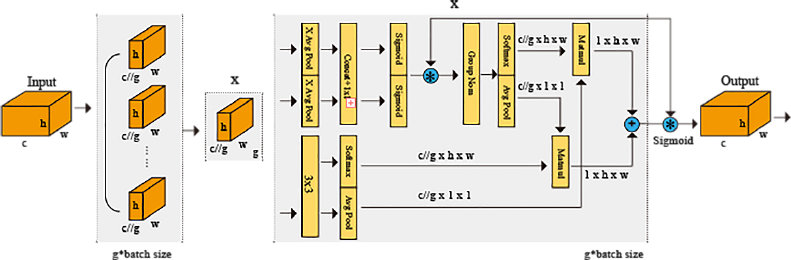}
\caption{EMA attention structure. ``g'' denotes channel groups, ``X Avg Pool'' horizontal pooling, and ``Y Avg Pool'' vertical pooling.}
\label{fig:fig4}
\end{figure}

An Exponential Moving Average (EMA) attention mechanism \cite{17} is incorporated into the later backbone layers (Fig.~\ref{fig:fig4}). By reweighting feature maps with pooled spatial statistics, EMA highlights informative regions and suppresses background noise. This improves the detection of small or occluded targets and enhances temporal consistency across frames, strengthening robustness in complex scenes.

\subsubsection{Pruning techniques}
\label{sssec:subsubhead3}
Channel pruning based on GroupNorm Importance \cite{42} is applied to further compress the model. The importance score of a channel $c$ is defined as:
\begin{align}
\label{eq:eq1}
I_{c} = \|W_{\rm c}\|_{2} = \sqrt{\sum_{i = 1}^{n}\omega_{c,i}^{2}},
\end{align}
where $W_{c}$ represents the channel weights and $n$ is the number of elements. Channels with the lowest scores are removed under a preset ratio, and the model is fine-tuned to recover performance. This process reduces parameters and FLOPs while preserving detection accuracy.

\subsection{Methods and principles of the combined model}
\label{ssec:subhead8}

\subsubsection{Overview of the TSN model}
\label{sssec:subsubhead4}
\begin{figure}[htb]
\centering
\includegraphics[width=0.9\linewidth]{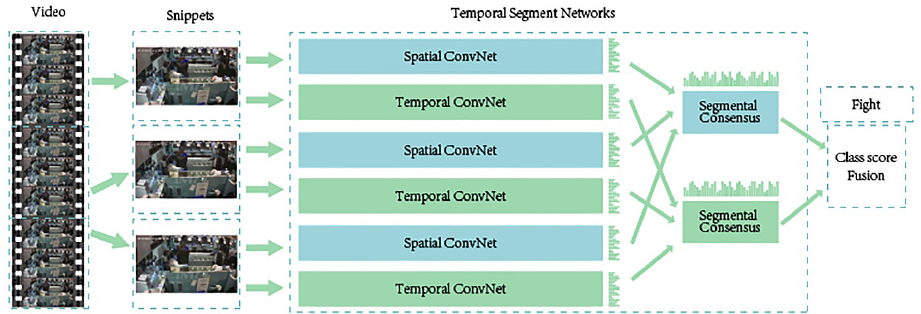}
\caption{Architecture of the TSN model.}
\label{fig:fig5}
\end{figure}

Temporal Segment Networks (TSNs) \cite{15} adopt a segment-based sampling strategy to capture long-term temporal dependencies in videos. Instead of processing entire sequences with recurrent models like LSTMs or applying costly 3D convolutions \cite{23}, TSNs divide a video into $K$ segments and sample one frame or short snippet from each, thereby achieving efficient temporal modeling with substantially lower FLOPs. A schematic overview is shown in Fig.~\ref{fig:fig5}.

\subsubsection{Integration with GE\mbox{-}YOLOv8}
\label{sssec:subsubhead5}

\textbf{(1) Object detection.}  
GE\mbox{-}YOLOv8 detects human instances in each frame, producing bounding boxes
\begin{align}
B_{t} = \{(x_{t,i}^{1}, y_{t,i}^{1}, x_{t,i}^{2}, y_{t,i}^{2}) \mid i = 1, \dots, N_{t}\},
\end{align}
where $N_{t}$ is the number of detected objects at frame $t$.

\textbf{(2) Region cropping.}  
Detected boxes are used to crop human regions
\begin{align}
I_{t,i} = [y_{t,i}^{1}, y_{t,i}^{2}, x_{t,i}^{1}, x_{t,i}^{2}],
\end{align}
thereby reducing background interference and focusing the subsequent temporal analysis.

\textbf{(3) Temporal modeling.}  
Cropped regions are fed into TSN, which samples $K$ segments to produce features
\begin{align}
{\rm F} = \{f_{1}, f_{2}, \dots, f_{K}\}, \quad
{\rm G} = {\rm TSN}({\rm F}).
\end{align}
The global feature ${\rm G}$ is then classified into violent or non-violent behavior:
\begin{align}
P_{\mathrm{violence}} = \sigma(WG+b),
\end{align}
where $W$ and $b$ denote the classifier parameters.

This pipeline enables precise spatial localization with GE\mbox{-}YOLOv8 and efficient temporal reasoning with TSN, forming the core of the proposed \textbf{Vi-SAFE} framework for violence recognition.

\section{Experiment}
\label{sec:experiment}

\subsection{Experiment Setting}
\label{ssec:subhead9}
The experimental platform is based on the ubuntu18.04 OS operating system, powered by an AMD EPYC 9754 CPU and an NVIDIA RTX 4090D graphics card with 24~GB of video memory. The environment is configured with Python 3.8, Torch 2.2.1, and CUDA 11.1.

\subsection{Dataset}
\label{ssec:subhead10}
We evaluated the proposed GE\mbox{-}YOLOv8 model and its TSN-integrated variant on two publicly available datasets. The pedestrian detection dataset, containing 1,539 images captured under diverse lighting conditions and crowd densities, was used to assess the model’s generalization ability. The RWF-2000 dataset, comprising 2,000 video clips equally divided into violent and non-violent categories, served as the primary benchmark. Both datasets were split into training and testing subsets in an 8:2 ratio. For GE\mbox{-}YOLOv8, performance was measured using Recall, mean Average Precision (mAP), parameter count, and GFLOPs, while the combined GE\mbox{-}YOLOv8+TSN model was evaluated using Accuracy (ACC).

\subsection{Analysis of GE\mbox{-}YOLOv8 Experimental Results}
\label{ssec:subhead11}

We evaluated the improved YOLOv8s modules through ablation studies on backbone substitution, attention integration, and pruning, each averaged over five runs.

\textbf{Backbone substitution.}  
Table~\ref{tab:tab1} compares lightweight backbones. GhostNetV3 reduces parameters by 53.2\% $\pm$ 1.2\% and GFLOPs by 56.7\% $\pm$ 1.5\% compared with YOLOv8s ($p<0.01$), while maintaining mAP 0.732, significantly outperforming MobileNetV4 and EfficientNetV2 ($p<0.05$).

\textbf{Attention integration.}  
Table~\ref{tab:tab2} shows EMA achieved the best balance (ACC 0.737, Recall 0.696), outperforming CBAM and CoordAtt ($p<0.05$). ANOVA confirmed significant differences across methods ($p<0.01$).

\textbf{Ablation results.}  
Table~\ref{tab:tab3} highlights that GhostNetV3+EMA achieved mAP 0.737 while halving parameters and FLOPs. Wilcoxon signed-rank test confirmed significance ($p<0.05$).

\textbf{Pruning.}  
Table~\ref{tab:tab4} shows pruning reduced parameters by 40.4\% and GFLOPs by 42.9\%, with minor decreases in Recall (\mbox{-}0.039) and mAP (\mbox{-}0.035). Differences were not significant ($p>0.05$), confirming effective complexity reduction.

\textbf{Overall.}  
GhostNetV3 and EMA enhance accuracy while reducing complexity, and pruning boosts efficiency, making GE\mbox{-}YOLOv8 suitable for real-time edge deployment.
\begin{table*}[t]
  \centering
  \begin{minipage}[t]{0.495\textwidth}
    \centering
    \captionof{table}{Backbone comparison on YOLOv8s.}
    \label{tab:tab1}
    \small
    \setlength{\tabcolsep}{4pt}
    \begin{tabular}{@{}lcccc@{}}
        \toprule
        \textbf{Model} & \textbf{Recall} & \textbf{mAP} & \textbf{Params} & \textbf{GFLOPs} \\
        \midrule
        YOLOv8s & 0.709 & 0.736 & 11.13\,M & 28.4 \\
        MobileNetV4 & 0.655 & 0.708 & 10.63\,M & 33.9 \\
        EfficientNetV2 & 0.657 & 0.715 & 8.79\,M & 8.3 \\
        GhostNetV3 & 0.695 & 0.732 & 5.92\,M & 16.1 \\
        \bottomrule
    \end{tabular}
    
    \vspace{1.5em}

    \captionof{table}{Attention mechanism comparison.}
    \label{tab:tab2}
    \small
    \setlength{\tabcolsep}{3pt}
    \begin{tabular}{@{}lcccc@{}}
        \toprule
        \textbf{Model} & \textbf{Recall} & \textbf{mAP} & \textbf{Params} & \textbf{GFLOPs} \\
        \midrule
        YOLOv8s & 0.709 & 0.736 & 11.13\,M & 28.4 \\
        GhostNetV3 & 0.655 & 0.708 & 10.63\,M & 16.1 \\
        GhostNetV3-CBAM & 0.670 & 0.728 & 6.18\,M & 16.3 \\
        GhostNetV3-CoordAtt & 0.675 & 0.727 & 5.95\,M & 16.1 \\
        GhostNetV3-EMA & 0.696 & 0.737 & 5.92\,M & 16.1 \\
        \bottomrule
    \end{tabular}
  \end{minipage}\hfill
  \begin{minipage}[t]{0.495\textwidth}
    \centering
    \setcounter{table}{3}
    \captionof{table}{Comparison of pruning experiments.}
    \label{tab:tab4}
    \small
    \setlength{\tabcolsep}{4pt}
    \begin{tabular}{@{}lcccc@{}}
        \toprule
        \textbf{Model} & \textbf{Recall} & \textbf{mAP} & \textbf{Params} & \textbf{GFLOPs} \\
        \midrule
        YOLOv8s & 0.709 & 0.736 & 11.13\,M & 28.4 \\
        GE-YOLOv8 (pre) & 0.696 & 0.737 & 5.92\,M & 16.1 \\
        GE-YOLOv8 (post) & 0.659 & 0.702 & 3.53\,M & 9.2 \\
        \bottomrule
    \end{tabular}
    
    \vspace{1.5em}

    \setcounter{table}{4}
    \captionof{table}{Recognition accuracy on RWF-2000 dataset.}
    \label{tab:tab5}
    \small
    \begin{tabular}{@{}lcc@{}}
        \toprule
        \textbf{Model} & \textbf{Core Operator} & \textbf{ACC} \\
        \midrule
        TSN(Baseline) & 2D CNN & 0.770 \\
        C3D \cite{42} & 3D CNN & 0.828 \\
        U-Net $+$ LSTM \cite{49} & 2D CNN $+$ LSTM & 0.820 \\
        ConvLSTM \cite{5} & 2D CNN $+$ LSTM & 0.770 \\
        TL \cite{50} & 3D CNN & 0.850 \\
        Openpose $+$ ST-GCN \cite{51} & 2D CNN $+$ GCN & 0.878 \\
        Veltmeijer et al.\cite{2025Real} & 3D CNN & 0.872 \\
        \textbf{Ours (Vi-SAFE)} & 2D CNN $+$ 2D CNN & \textbf{0.880} \\
        \bottomrule
    \end{tabular}
  \end{minipage}

  \vspace{2em}

  \setcounter{table}{2}
  \centering
  \caption{Ablation experiments of GE-YOLOv8.}
  \label{tab:tab3}
  \small
  \begin{tabular*}{\textwidth}{@{\extracolsep{\fill}}ccccccc}
      \toprule
      \textbf{YOLOv8s} & \textbf{GhostNetV3} & \textbf{EMA} & \textbf{Recall} & \textbf{mAP} & \textbf{Params} & \textbf{GFLOPs} \\
      \midrule
      $\surd$ & & & 0.709 & 0.736 & 11.13\,M & 28.4 \\
      $\surd$ & & $\surd$ & 0.696 & 0.741 & 11.13\,M & 28.5 \\
      $\surd$ & $\surd$ & & 0.695 & 0.732 & 5.92\,M & 16.1 \\
      $\surd$ & $\surd$ & $\surd$ & 0.696 & 0.737 & 5.92\,M & 16.1 \\
      \bottomrule
  \end{tabular*}
\end{table*}

\subsection{Experimental Results and Analysis of Vi-SAFE}
\label{ssec:subhead12}

We evaluated the proposed \textbf{Vi-SAFE} framework, which integrates GE\mbox{-}YOLOv8s with TSN, on the RWF-2000 dataset. Accuracy (ACC) was used as the evaluation metric, defined as the proportion of correctly predicted videos relative to the total number of test videos. All results were averaged over five independent runs.

As shown in Table~\ref{tab:tab5}, \textbf{Vi-SAFE} achieved an ACC of 0.880 $\pm$ 0.007, outperforming conventional baselines such as TSN (0.770), U-Net+LSTM (0.820), and C3D (0.828), and slightly surpassing the Openpose+ST-GCN method (0.878 $\pm$ 0.006). Statistical tests confirmed that these improvements were significant ($p<0.01$, one-way ANOVA). The superior performance stems from the complementary design of GE\mbox{-}YOLOv8 and TSN: the former provides efficient and precise spatial feature extraction, while the latter captures long-range temporal dynamics. Unlike 3D CNN approaches, which incur high computational costs, or 2D CNN+LSTM methods, which struggle with temporal consistency, the dual 2D CNN structure in Vi-SAFE delivers higher accuracy with lower complexity, making it suitable for edge deployment. Furthermore, the modular architecture of Vi-SAFE ensures scalability, allowing the integration of additional modules for broader applications in public safety surveillance.

\section{Conclusions}
\label{sec:conclusions}

This study introduces Vi-SAFE, a novel spatiotemporal framework that combines GE\mbox{-}YOLOv8 with TSN for the detection of violent behavior in videos. With pruning optimization, GE\mbox{-}YOLOv8 maintains high accuracy while significantly reducing parameters and GFLOPs. Integrated with TSN, Vi-SAFE achieves an ACC of 0.88 on the RWF-2000 dataset, surpassing existing mainstream methods. The framework demonstrates strong scalability through its modular design, allowing easy adaptation to diverse application scenarios and the integration of additional modules. While Vi-SAFE exhibits robustness and efficiency, future work will focus on improving generalization in complex environments by evaluating more diverse datasets and exploring advanced optimization strategies.

\newpage
\bibliographystyle{IEEEbib}
\bibliography{refs} 

\end{document}